\documentclass[letterpaper]{article} 
\usepackage{aaai24}  
\usepackage{times}  
\usepackage{helvet}  
\usepackage{courier}  
\usepackage[hyphens]{url}  
\usepackage{graphicx} 
\usepackage{natbib}  
\usepackage{caption} 
\usepackage[utf8]{inputenc}
\usepackage{algorithm,algorithmicx,algpseudocode}
\algrenewcommand\algorithmiccomment[1]{\hfill #1}

\usepackage{xcolor}
\usepackage{textcomp}
\usepackage{booktabs}
\usepackage{amsfonts}
\usepackage{multirow}
\usepackage{diagbox}
\usepackage{colortbl}

\usepackage{newfloat}
\usepackage{listings}
\DeclareCaptionStyle{ruled}{labelfont=normalfont,labelsep=colon,strut=off} 
\floatstyle{ruled}
\newfloat{listing}{tb}{lst}{}
\floatname{listing}{Listing}
\title{Transformer Multivariate Forecasting: Less is More?}
\author {
    Jingjing Xu\textsuperscript{\rm 1},
    Caesar Wu\textsuperscript{\rm 1, 3},
    Yuan-Fang Li\textsuperscript{\rm 2},
    Pascal Bouvry\textsuperscript{\rm 1, 3}
}
\affiliations {
    \textsuperscript{\rm 1} University of Luxembourg/FSTM\\
    \textsuperscript{\rm 2} Monash University\\
    \textsuperscript{\rm 3} Interdisciplinary Centre for Security, Reliability and Trust (SnT)\\
    jingjing.xu@uni.lu, caesar.wu@uni.lu, yuanfang.li@monash.edu, pascal.bouvry@uni.lu
}

\begin{document}

\maketitle

\begin{abstract}
In the domain of multivariate forecasting, transformer models stand out as powerful apparatus, displaying exceptional capabilities in handling messy datasets from real-world contexts. However, the inherent complexity of these datasets, characterized by numerous variables and lengthy temporal sequences, poses challenges, including increased noise and extended model runtime. This paper focuses on reducing redundant information to elevate forecasting accuracy while optimizing runtime efficiency. We propose a novel transformer forecasting framework enhanced by Principal Component Analysis (PCA) to tackle this challenge. The framework is evaluated by five state-of-the-art (SOTA) models and four diverse real-world datasets. Our experimental results demonstrate the framework's ability to minimize prediction errors across all models and datasets while significantly reducing runtime. From the model perspective, one of the PCA-enhanced models: PCA+Crossformer, reduces mean square errors (MSE) by 33.3\% and decreases runtime by 49.2\% on average. From the dataset perspective, the framework delivers 14.3\% MSE and 76.6\% runtime reduction on Electricity datasets, as well as 4.8\% MSE and 86.9\% runtime reduction on Traffic datasets. This study aims to advance various SOTA models and enhance transformer-based time series forecasting for intricate data. Code is available at: \url{https://github.com/jingjing-unilu/PCA_Transformer}
\end{abstract}

\section{Introduction}
\label{Introduction}
Sequence modeling proves highly effective in capturing patterns within sequential data types like languages, time series, and biological data. Various forms of recurrent neural networks (RNNs)~\cite{schuster1997bidirectional} play a vital role in this modeling. Unlike traditional fully connected neural networks (FCNs), which struggle to share features across different data points or locations, standard RNNs overcome this limitation by incorporating input from the previous time step, facilitating the connection of features across various locations within the sequence. 

However, standard RNNs face barriers in capturing long-term dependencies due to issues like vanishing and exploding gradients. Proposed solutions, including the integration of Rectified Linear Unit (ReLU) activation functions, weight initialization with identity matrices, and the use of gates within RNNs, aim to mitigate these challenges. Other approaches, such as Gated Recurrent Units (GRU)~\cite{cho2014learning} and Long Short-Term Memory (LSTM)~\cite{hochreiter1997long}, intend to leverage gate cells to retain long-term dependencies. Nonetheless, challenges persist, particularly in unidirectional processing during training.
\begin{figure}
\begin{center}
\includegraphics[width=6.5cm]{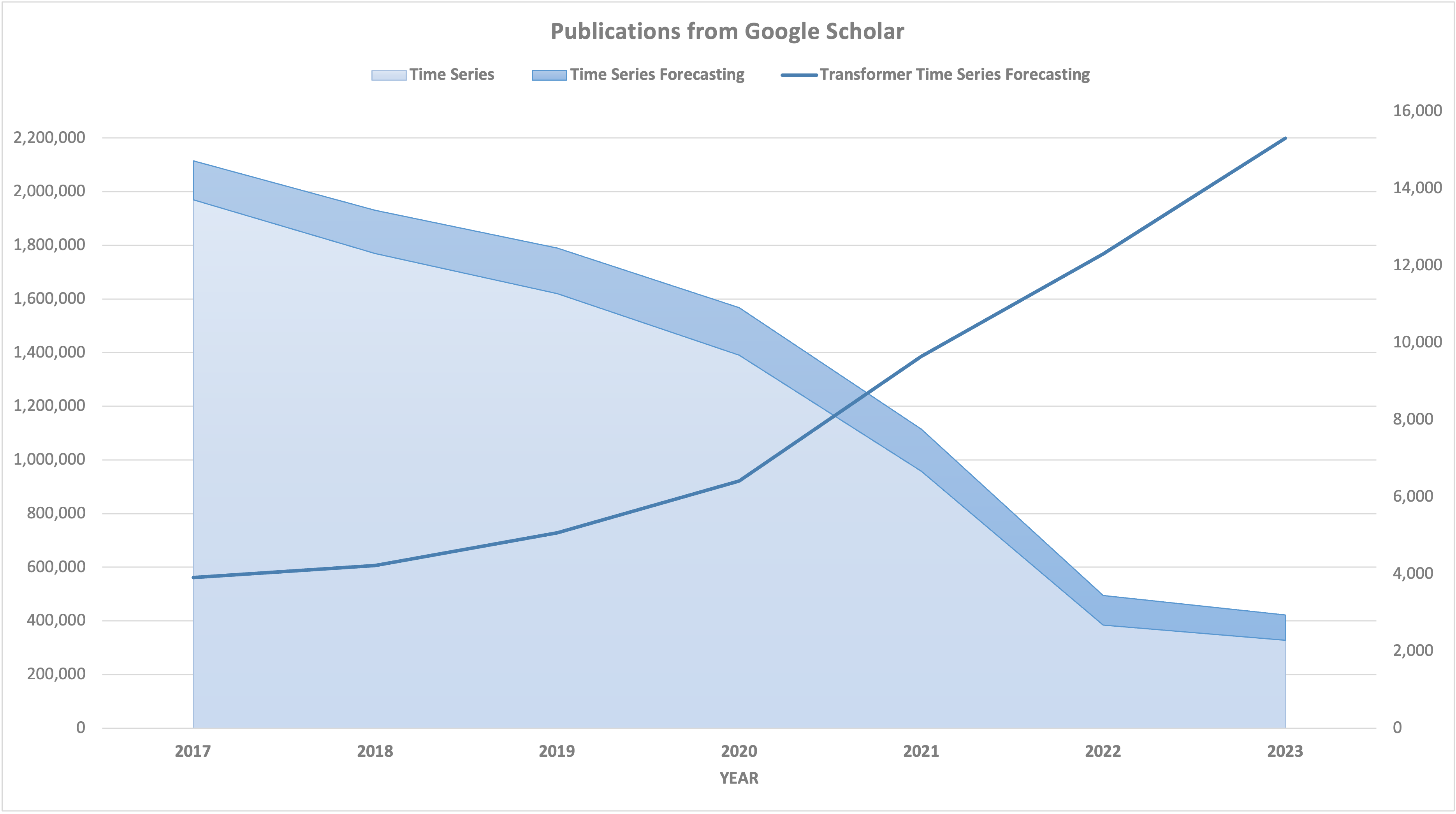}
\end{center} 
\caption{Trend of transformer in time series forecasting topic from 2017.}
\label{T_in_TS}
\end{figure}
In response to this limitation, bidirectional RNN structures have been introduced, enabling the processing of both forward and reverse-directional information within a sequence. Despite advancements, these models still face challenges attributed to a lack of parallelism. The Transformer architecture~\cite{vaswani2017attention} offers a solution by concurrently processing sequence data through an encoder-decoder architecture and employing attention mechanisms on each word. This approach enhances both efficiency and performance. Transformers prove particularly adept at handling long-term sequence data, such as lengthy sentences in Natural Language Processing (NLP)~\cite{gillioz2020overview}, as well as image and video data in the Computer Vision (CV)~\cite{han2022survey} domain. This characteristic of the transformer architecture in long-time series forecasting can also be applied in other domains. 

The landscape of time series forecasting models spans from classic auto-regressive-moving-average (ARMA) models~\cite{whittle1951hypothesis} to the era of deep learning. The Transformer, initially crafted for sequential data, emerges as an appealing solution for studying its application in time series forecasting tasks. Figure~\ref{T_in_TS} shows the trend of Google Scholar's publications on Transformer techniques since 2017. This paper specifically aims for long-term multivariate forecasting, seeking to predict the future based on multiple variables using transformer models. 

While multivariate datasets provide richer information and insightful patterns for constructing prediction models, they are inherently complex and pose challenges such as sensitivity to errors and increased computational demands. Excessive runtimes during training and testing phases contribute to significant energy consumption and a large carbon footprint. Hence, the necessity arises for performing multivariate analysis, dimensionality reduction, and feature extraction to facilitate model implementation. Many researchers have made significant contributions by offering different approaches on dimensionality reduction, including Principal Components Analysis (PCA)~\cite{pearson1901liii,hotelling1933analysis}, Linear Discriminant Analysis (LDA), and t-Distributed Stochastic Neighbor Embedding (t-SNE)~\cite{anowar2021conceptual}. However, the existing approaches to dimensionality reduction in datasets for transformer models remain insufficient. This paper addresses this gap through a series of intensive experiments. This study introduces a novel framework for multivariate analysis, emphasizing dimension reduction in the context of transformer-based multivariate time series forecasting from a dataset perspective. 
\subsection{Contributions}
The primary contributions of this paper are as follows:
\begin{enumerate}
    \item We present a benchmark test on transformer-based multivariate long-term forecasting with PCA dimension reduction. The comprehensive benchmark test is a particular scheme to evaluate the performance of multivariate long-term forecasting.
    \item The work demonstrates that the proposed framework can significantly enhance long-term forecasting performance across five SOTA representative models and four real-world datasets. 
    \item The comprehensive study also provides detailed and interpretable insights for model transparency, which adds the value of the interpretability and explainability of multivariate analysis underpinned by PCA techniques.
\end{enumerate}
The rest of this paper is organized as follows: Section 2 is the related work. Section 3 is the experiments part, which defines datasets, experimental formulation, experimental environment and configuration, as well as the framework construction. Section 4 shows the experimental results with analysis. Section 5 presents the conclusion and future works.  
\section{Related Work}
\label{Background}
\subsection{Transformer in Time Series Forecasting}
Transformer models display excellent performance due to its ability to capture long-range dependencies in sequential tokens. Consequently, the transformer model emerges as a good option for modelling time series problems. Several transformer-based forecasting models have been developed to address specific challenges in time series forecasting. Transformer-based forecasting models can be organized into different categories based on network modification criteria~\cite{wen2022transformers,liu2023itransformer}. This research adopts the classification method articulated by~\cite{liu2023itransformer}. Table~\ref{tran_cate} highlights four types of transformer models based on modified components and architecture.
\begin{table}[ht]
\centering
\resizebox{\linewidth}{!}{
\begin{tabular}{c|c|c}
\toprule
\diagbox{\bf Modi. Archi.}{\bf Modi. Compo.}    & \textbf{Yes}                           & \textbf{No}       \\        
\hline
\multirow{2}{*}{\bf Yes}        & \multirow{2}{*}{Crossformer}  &  \multirow{2}{*}{iTransformer}  \\
                            &                               &                                 \\
\hline
\multirow{2}{*}{\bf No}         & \multirow{2}{*}{Autoformer}   & PatchTST \\
                            &                               & Non-stationary Transformer \\
\bottomrule                           
\end{tabular}
}
\caption{Four categories of transformer-based forecasting models. Modi. Archi.: Modified Architecture; Modi. Compo.: Modified Components.}
\label{tran_cate}
\end{table}

Informer~\cite{zhou2021informer} and Autoformer~\cite{wu2021autoformer} focuses on adapting components for temporal long sequences. The Crossformer model~\cite{zhang2022crossformer} presents its capability to preserve time and dimension information while effectively capturing cross-time and cross-dimension dependencies, enhancing its performance in multivariate time series forecasting. The Non-stationary Transformer~\cite{liu2022non}), on the other hand, unifies input, converts output, and solves the over-stationary problem to forecast results. PatchTST~\cite{nie2022time} utilizes patching and channel-independent architecture. It facilitates the model to capture local semantic information and longer lookback windows. iTransformer~\cite{liu2023itransformer} inverts the structure of the transformer model without modify components to enhance the performance of forecasting. It provides an alternative architecture for time series forecasting. Additionally, many recent transformer-based models such as Scaleformer~\cite{shabani2022scaleformer}, TDformer~\cite{zhang2022first}, GCformer~\cite{zhao2023gcformer}, PDTrans~\cite{tong2023probabilistic} etc, offering diverse approaches to time series forecasting. In addition,~\cite{wu2024trustworthy,wu2024strategic} introduces credit default swap (CDS) dataset for testing transformer models. A comprehensive survey by~\cite{wen2022transformers} explores transformer models for time series forecasting, which provides some valuable insights. Besides transformer models, researchers also adopt other techniques to address time series forecasting, such as graph representation learning~\cite{jin2022multivariate} and Large Language Models (LLMs)~\cite{jin2023time}. 
\subsection{PCA}
\begin{figure*}
\begin{center}
\includegraphics[width=7in]{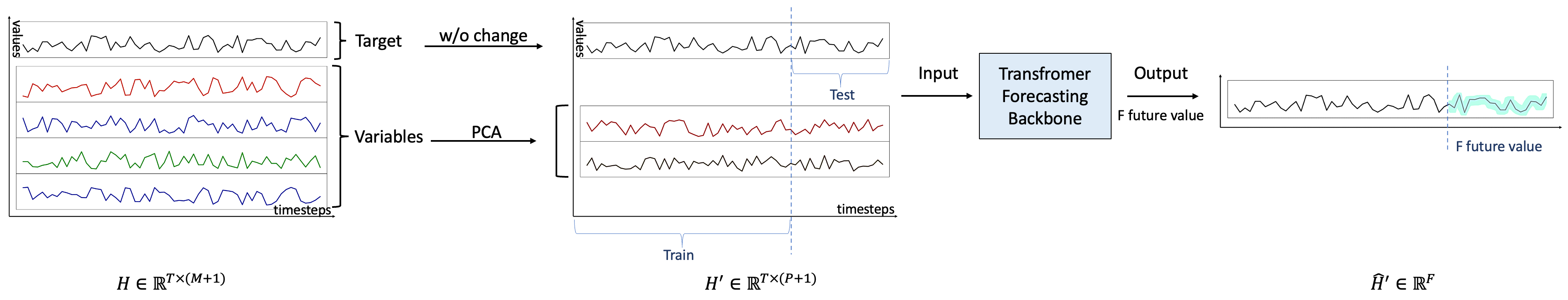}
\end{center} 
\caption{Structure overview of the PCA-enhanced transformer forecasting framework.}
\label{structure_overview}
\end{figure*}
\begin{figure}
\centering
\begin{center}
\includegraphics[width=6.5cm]{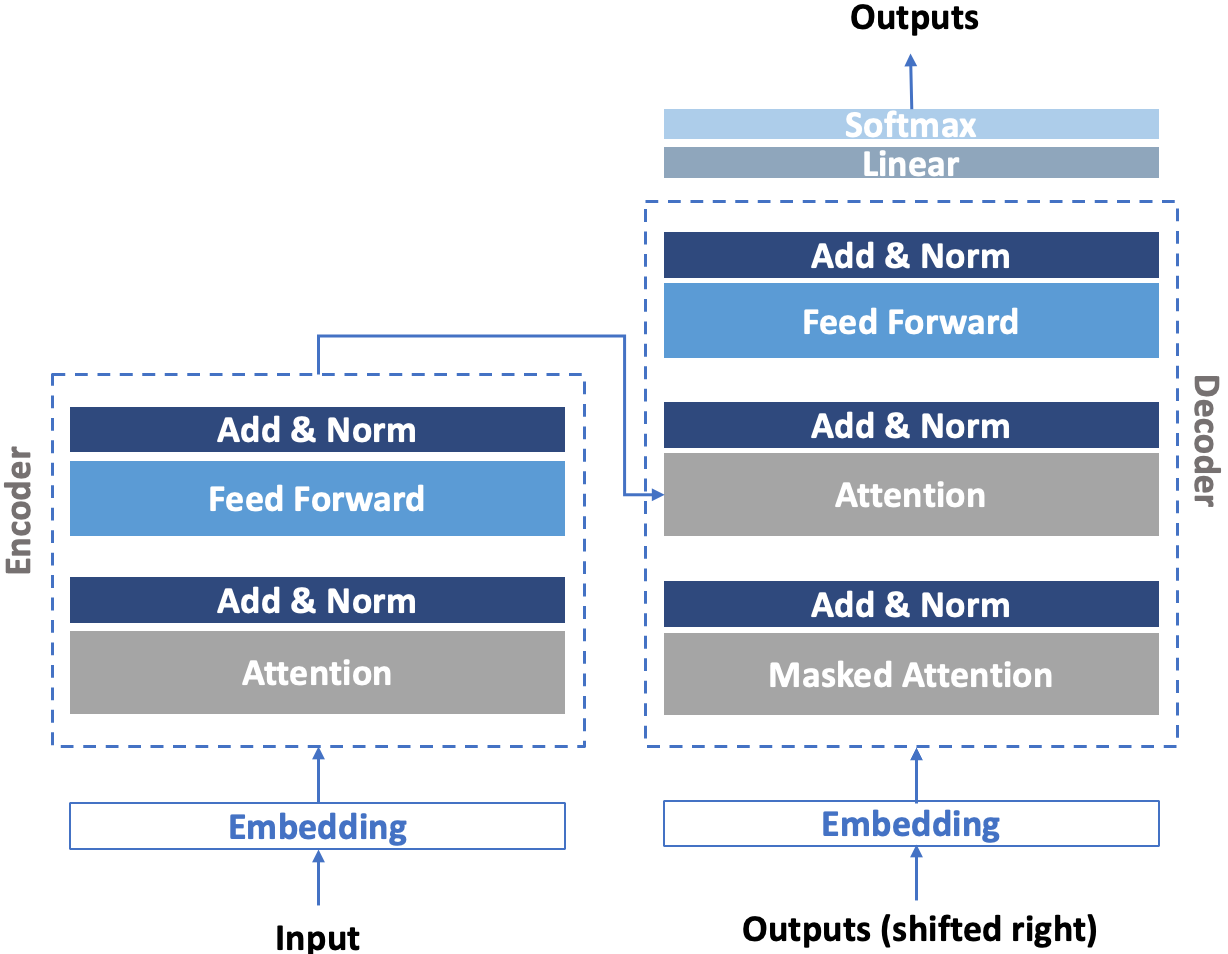}
\end{center} 
\caption{Simplified vanilla transformer architecture with three main components (Attention, Add\&Norm, FeedForward).}
\label{Tran_arch}
\end{figure}
From a data dimensionality reduction perspective, PCA has a distinct capacity to reduce dimensions in an easily interpretable way while preserving the essential information contained in the data. The roots of the PCA technique can be traced back to~\cite{pearson1901liii} and~\cite{hotelling1933analysis}. PCA has gained widespread acceptance, particularly in the era of high-dimensional big data, such as image, text, various stock market data~\cite{xie2019principal} and hospital patient data~\cite{kutcher2013principal} etc, PCA has widespread adoption. The methodology has evolved into several versions, such as Sparse PCA~\cite{hotelling1933analysis,merola2015least}, Nonlinear PCA~\cite{hastie1989principal}, Kernel PCA~\cite{scholkopf1998nonlinear} and Robust PCA~\cite{kriegel2008general}. It has become a fundamental tool for unraveling and interpreting complex datasets across varied domains.
\subsection{PCA+Transformer in Time Series Forecasting}
Several studies have explored the integration of PCA with transformer in time series forecasting, and they focus on different perspectives. For instance, the work by~\cite{an2022novel} centers on combining PCA with Informer model for fault detection and prediction in nuclear power valves. However, this work suffers from shortage of experimental implementation across various transformer models and datasets. Alongside this development, PCA has gained popularity for its role in model interpretation. In recent studies~\cite{madane2022transformer} and~\cite{li2022tts}, PCA has been applied to assess model performance, specifically in examining similarities between synthetic and real data. Researchers~\cite{zhou2023one} present a distinctive viewpoint by connecting self-attention with PCA to illustrate the functioning of transformer models. Data analysts~\cite{jin2022time} adopt t-distributed stochastic neighbour Embedding (t-SNE) method (an unsupervised nonlinear dimensionality reduction technique, similar to the PCA) to explore the operational patterns of the model. Other investigators~\cite{pandey2023multi} have introduced kernel PCA models for multivariate time series forecasting but have not applied them to various transformer-based forecasting models. The limitations of dimensionality reduction techniques remain when applied to datasets intended for transformer-based models. To address this gap, we propose the PCA-enhanced transformer-based time series forecasting model.



\section{Experiments}
\subsection{Dataset}
To evaluate all transformer models, we adopt four long-term time series datasets widely accepted as standard data sources. They are ETTh1, Weather, Traffic, and Electricity. The targeted or predicted variables for these datasets are oil temperature, wet bulb, 862th sensor, and 321th client, respectively. Furthermore, these four datasets contain different ranges of variables, varying from 7 to 862 with large time steps. Having dataset with large variables and long time steps can ensure stable results with less over-fitting. Table~\ref{ds_bench} highlights a summary of information about four datasets. A detailed description of variables for these datasets can be found in~\cite{zhou2021informer,wu2021autoformer}.
\begin{table}[!ht]
    \centering
    \begin{tabular}{c|c|c|c|c}
     \toprule
        Datasets & ETTh1 & Weather & Electricity & Traffic \\ 
        \hline
        Variables & 7 & 21 & 321 & 862 \\ 
        \hline
        Timesteps & 17420 & 52696 & 26304 & 17544 \\
    \bottomrule
    \end{tabular}
    \caption{Summary of information about four datasets (targets are counted into variables).}
    \label{ds_bench}
\end{table}
\subsection{Experimental Formulation}
We aim to solve the following problem: given past observations of time series $ H = (x^{1}, x^{2}, ..., x^{M}, y)\in \mathbb{R}^{T\times (M+1)} $ where the $M$ represents $M$ variables, the $y$ denotes the \textit{target} and the $T$ is the length of time series. The time series of the $m$-th variable is denoted as $x^{m} = (x_{1}^{m}, x_{2}^{m}, ..., x_{T}^{m})^\top\in \mathbb{R}^{T}$. The time series of the \textit{target} can be denoted as $y = (y_{1}, y_{2}, ..., y_{T})^\top\in \mathbb{R}^{T}$. We aim to forecast $F$ future values of the \textit{target} $(y_{T+1}, y_{T+2}, ..., y_{T+F})^\top$.
\begin{table*}[!ht]
\centering
\resizebox{\textwidth}{!}{
\begin{tabular}{c|c|c|rr|rr|rr|rr|rr|rr}
\toprule
\multirow{2}{*}{\textbf{Dataset}}                       & \textbf{Variables}       & \textbf{PCA} & \multicolumn{2}{c|}{\textbf{PCA+PatchTST}}                             & \multicolumn{2}{c|}{\textbf{PCA+Crossformer}}                          & \multicolumn{2}{c|}{\textbf{PCA+Autoformer}}                           & \multicolumn{2}{c|}{\textbf{PCA+N.S. Trans.}}           & \multicolumn{2}{c|}{\textbf{PCA+iTransformer}}         & \multicolumn{2}{c}{\textbf{PCA+Transformer}}                          \\ 
 & \textbf{w/o Target}       & \textbf{Components} & \multicolumn{2}{c|}{(\bf MSE $\mid$ \bf MAE)}    & \multicolumn{2}{c|}{(\bf MSE $\mid$ \bf MAE)}  & \multicolumn{2}{c|}{(\bf MSE $\mid$ \bf MAE)}   & \multicolumn{2}{c|}{(\bf MSE $\mid$ \bf MAE)}       & \multicolumn{2}{c|}{(\bf MSE $\mid$ \bf MAE)}  & \multicolumn{2}{c}{(\bf MSE $\mid$ \bf MAE)}        
 \\ 
\midrule
     &                                & \textbf{2}              & \cellcolor[HTML]{D4E6F1}{\bf 0.05561}   & 0.17868       & 0.19371 & 0.38307      & \cellcolor[HTML]{D4E6F1}{\bf 0.08919} & \bf 0.22736   
     & \cellcolor[HTML]{D4E6F1}{\bf 0.07144}   & \bf 0.20484       & 0.05736 & 0.18378                    & 0.85307  & 0.88175               \\ 
     &                                & \textbf{4}              & {0.05671} & 0.18052                     & {\cellcolor[HTML]{D4E6F1}{\bf 0.16091}}  & {\bf 0.33508}         & {0.10550} & 0.25530                          & {0.08048} & 0.21217               & {\cellcolor[HTML]{D4E6F1}{\bf 0.05637}}   & {\bf 0.18241}                      & {1.17205} & 1.01673               \\ 
\multirow{-3}{*}{\textbf{ETTh1}}        & \multirow{-3}{*}{\textbf{6}}   & \textbf{w/o PCA}        & {0.05575} & {\bf 0.17838}        & {0.36994} & 0.55491               & {0.11395}   & 0.25875               & {0.07814}  & 0.21079               & {0.05662} & 0.18278                  & {\cellcolor[HTML]{D4E6F1}{\bf 0.72492}}   & {\bf 0.79393}  
\\ 
\midrule
     &                                & \textbf{2}              & {0.0013408}  & 0.0267667         & {0.0066745}  & 0.0669198                                & {0.0390251} & 0.0985756                         & {\cellcolor[HTML]{D4E6F1}{\bf 0.0012783}} & {\bf 0.0262055}           & {\cellcolor[HTML]{D4E6F1}{\bf 0.0012391}} & {\bf 0.0258336}                    & {0.0074935} & 0.0697642    
\\ 
     &                                & \textbf{5}              & {0.0013426} & 0.0266463            & {0.0039031}  & 0.0495619                                & {0.0086512}  & 0.0701804             & {0.0013453} & 0.0270548             & {0.0012655} & 0.0262108                       & {0.0031227}  & 0.0433567                   
\\ 
     &                                & \textbf{10}             & {0.0013277}  & 0.0265256             & {0.0044438} & 0.0535568                           & {0.0114851} & 0.0851401             & {0.0015487}  & 0.0288031             & {0.0012676} & 0.0260793       & {0.0051427}  & 0.0554222        
\\
     &                                & \textbf{15}             & {0.0013124}  & 0.0266787            & {\cellcolor[HTML]{D4E6F1}{\bf 0.0019383}} & {\bf 0.0324117}                      & {\cellcolor[HTML]{D4E6F1}{\bf 0.0072318}} & {\bf 0.0679583}                       & {0.0018783} & 0.0313342             & {0.0013160}  & 0.0265606                                & {0.0040705}  & 0.0517972                     
\\ 
\multirow{-5}{*}{\textbf{Weather}} & \multirow{-5}{*}{\textbf{20}}  & \textbf{w/o PCA}        & {\cellcolor[HTML]{D4E6F1}{\bf 0.0013116}} & {\bf 0.0263536}           & {0.0045753}& 0.0548066                     & {0.0079332}  & 0.0704268             & {0.0016366} & 0.0294900             & {0.0013511}  & 0.0271509                                 & {\cellcolor[HTML]{D4E6F1}{\bf 0.0025455}} & {\bf 0.0372824} 
\\ 
\midrule
     &                                & \textbf{2}              & {\cellcolor[HTML]{D4E6F1}{\bf 0.30114}}   & {\bf 0.39454}        & {0.24117} & 0.35286                        & {0.41615} & 0.49454                       & {0.32548} & 0.42185               & {0.35119} & 0.43913& {\cellcolor[HTML]{D4E6F1}{\bf 0.31722}}   & {\bf 0.42333}   
\\ 
     &                                & \textbf{20}             & {0.30803}  & 0.40253          & {0.31382} & 0.39153                         & {0.39585}  & 0.47261                     & {0.35360} & 0.44082            & {0.31114} & 0.41341                                  & {0.44774}   & 0.50390         
\\ 
     &                                & \textbf{40}             & {0.31029}  & 0.40227               & {0.32395}  & 0.39478                     & {0.38023}  & 0.46607                    & {0.38404} & 0.44949               & {0.34803}  & 0.43267                        & {0.45457}  & 0.49402             
\\ 
     &                                & \textbf{80}             & {0.30792}  & 0.40005           & {\cellcolor[HTML]{D4E6F1}{\bf 0.23528}}   & {\bf 0.34291}         & {0.42425} & 0.48486& {0.31263} & 0.41109             & {0.28967} & 0.39682                                  & {0.36052} & 0.44804               
\\ 
     &                                & \textbf{160}            & {0.30951} & 0.39863   & {0.30753}& 0.38397                       & {0.38894}  & 0.47346                                 & {0.32938}  & 0.42110               & {0.31660} & 0.41921                         & {0.48044}   & 0.51270     
\\ 
     &                                & \textbf{240}            & {0.31747}   & {0.40354}                             & {0.26574}  & {0.36454}                        & {0.42832} & {0.49689}                               & {0.34589}  & {0.43446}                             & {\cellcolor[HTML]{D4E6F1}{\bf 0.25611}}   & {\bf 0.36751}         & {0.37116} & {0.45557}                             
\\ 
\multirow{-7}{*}{\textbf{Electricity}}  & \multirow{-7}{*}{\textbf{320}} & \textbf{w/o PCA}        & {0.30771} & 0.40156             & {0.27444}  & 0.36760                   & \cellcolor[HTML]{D4E6F1}{\bf 0.37565}  & {\bf 0.46435}                              & \cellcolor[HTML]{D4E6F1}{\bf 0.30231}  & {\bf 0.40977}              & {0.54722}  & 0.54924                                 & {0.39571} & 0.46279      
\\ 
\midrule
     &                                & \textbf{1}              & {0.17485}  & 0.25452        & \cellcolor[HTML]{D4E6F1}{\bf 0.14789} & {\bf 0.22577}                & {0.31848}  & 0.41374                                & {0.18998}   & {0.29156}                & {0.32595}  & 0.40816               & {0.28672}  & {0.35876}   
\\ 
     &                                & \textbf{2}              & {0.17714}  & 0.25728         & {0.16042} & 0.24255                & {0.29514}  & 0.39557                                & {\cellcolor[HTML]{D4E6F1}{\bf 0.17894}}   & {\bf 0.28501}                & {1.86223}  & 1.17306               & {\cellcolor[HTML]{D4E6F1}{\bf 0.28144}}   & {\bf 0.35329}   
\\ 
     &                                & \textbf{25}             & {\cellcolor[HTML]{D4E6F1}{\bf 0.16754}}   & 0.24795                    & {0.25004}  & 0.31726         & {0.27335} & 0.36999                               & {0.22374}  & 0.33921               & {1.31162} & 0.93039                          & {0.29579} & 0.37345                  
\\ 
     &                                & \textbf{50}             & {0.16907}  & 0.24932                     & {0.20874}   & 0.28413            & {\cellcolor[HTML]{D4E6F1}{\bf 0.24967}}   & {\bf 0.34481}   & {0.20833}  & 0.31271                             & {1.86220} & 1.17306                                  & {0.41403} & 0.43881  
\\ 
     &                                & \textbf{105}            & {0.17121} & 0.25329                        & {0.23546} & 0.29953               & {0.30357}   & 0.40072                  & {0.19733}  & 0.30020               & {1.86222}  & 1.17307                     & {0.33498}  & 0.40060        
\\
     &                                & \textbf{215}            & {0.17100}  & 0.24890                       & {0.16648} & 0.23886               & {0.37778}   & 0.44868                    & {0.23479}   & 0.33047               & {1.86224} & 1.17306                        & {0.35731}  & 0.41112       
\\ 
     &                                & \textbf{430}            & {0.17507}  & 0.25062& {0.17663}   & 0.25027                              & {0.29786}   & 0.40188               & {0.27937} & 0.38314               & {1.86222} & 1.17307             & {0.38011}  & 0.41905   
\\ 
     &                                & \textbf{645}            & {0.17123}   & 0.24950                                  & {0.21702}  & 0.28981                            & {0.31110} & 0.39311               & {0.37402}    & 0.45356               & {1.25570}   & 0.92052              & {0.33137}  & 0.39332                         
\\ 
\multirow{-9}{*}{\textbf{Traffic}}      & \multirow{-8}{*}{\textbf{861}} & \textbf{w/o PCA}        & {0.16831} & {\bf 0.24585}   & {{0.15540}}   & {0.22792}        & {0.30356}   & 0.40767               & {0.21742}  & 0.32678                                & {\cellcolor[HTML]{D4E6F1}{\bf 0.25060}}   & {\bf 0.35071}                      & {0.29690} & 0.35520                              
\\ 
\bottomrule
\end{tabular}
}
\caption{Accuracy results of transformer forecasting models (w/o PCA) and PCA-enhanced transformer forecasting models. Note: 1) w/o PCA: without PCA; 2) PCA+N.S. Trans.: PCA + Non-stationary Transformer.}
\label{acc_result_PCA_tran}
\end{table*}
\begin{table}[!ht]
\centering
\resizebox{\linewidth}{!}{
\begin{tabular}{c|c|c|c|c|c|c|c|c}
\toprule
\multirow{2}{*}{\textbf{Dataset}}   & \textbf{Variables}  & \textbf{PCA} & {\textbf{PCA+}}  & {\textbf{PCA+}}  & {\textbf{PCA+}}    & {\textbf{PCA+N.S.}}    & {\textbf{PCA+}}   & {\textbf{PCA+}}   \\ 
~ & \textbf{w/o Target}       & \textbf{Components} & {\textbf{PatchTST}}     & {\textbf{Crossformer}}  & {\textbf{Autoformer}}   & {\textbf{Transformer}}     & {\textbf{iTransformer}} & {\textbf{Transformer}}   
 \\ 
\midrule
     &                                & \textbf{2}              & {\cellcolor[HTML]{DDEBF7}\textbf{71}}                  & {241}               & \cellcolor[HTML]{DDEBF7}\textbf{310}                        & \cellcolor[HTML]{DDEBF7}\textbf{118}                          & 87  & 219 \\ 
     
     &                                & \textbf{4}              & {64}                & {\cellcolor[HTML]{DDEBF7}\textbf{246}}                 & 311                      & 141                        & \cellcolor[HTML]{DDEBF7}\textbf{84}    & 126 \\ 
\multirow{-3}{*}{\textbf{ETTh1}}        & \multirow{-3}{*}{\textbf{6}}   & \textbf{w/o PCA}        & {74}                & {252}               & {310} & {177}   & 88  & {\cellcolor[HTML]{DDEBF7}\textbf{126}}                 \\ \midrule
     &                                & \textbf{2}              & 231 & 1915& 686                      & \cellcolor[HTML]{DDEBF7}\textbf{415}                          & \cellcolor[HTML]{DDEBF7}\textbf{653}   & 375 \\ 
     &                                & \textbf{5}              & 241 & 1264& 690                      & 414                        & 538 & 377 \\ 
     &                                & \textbf{10}             & 250 & 1399& 701                      & 416                        & 603 & 379 \\ 
     &                                & \textbf{15}             & 283 & \cellcolor[HTML]{DDEBF7}\textbf{1224}  & \cellcolor[HTML]{DDEBF7}\textbf{689}                        & 418                        & 440 & 378 \\ 
\multirow{-5}{*}{\textbf{Weather}} & \multirow{-5}{*}{\textbf{20}}  & \textbf{w/o PCA}        & \cellcolor[HTML]{DDEBF7}\textbf{311}   & 1768& {685} & {417}   & 625 & {\cellcolor[HTML]{DDEBF7}\textbf{378}}                 \\ \midrule
     &                                & \textbf{2}              & \cellcolor[HTML]{DDEBF7}\textbf{221}   & { 1430}                                 & 975                      & 876                        & {200}                                  & \cellcolor[HTML]{DDEBF7}{\textbf{283}} \\ 
     &                                & \textbf{20}             & 246 & {883}                                  & 723                      & 1198                       & {209}                                  & {241}                                  \\ 
     &                                & \textbf{40}             & 352 & {904}                                  & 604                      & 1043                       & {207}                                  & {241}                                  \\ 
     &                                & \textbf{80}             & 579 & {\cellcolor[HTML]{DDEBF7} \textbf{925}} & 733                      & 887                        & { 205}                                  & {245}                                  \\ 
     &                                & \textbf{160}            & 1038& {1620}                                 & 748                      & 1394                       & { 216}                                  & {250}                                  \\ 
     &                                & \textbf{240}            & 1510& { 2847}                                 & 622                      & 1418                       & \cellcolor[HTML]{DDEBF7}{\textbf{380}} & {303}                                  \\ 
\multirow{-7}{*}{\textbf{Electricity}}  & \multirow{-7}{*}{\textbf{320}} & \textbf{w/o PCA}        & 1953              & 3960            & \cellcolor[HTML]{DDEBF7}\textbf{759}                        & \cellcolor[HTML]{DDEBF7}\textbf{1768}                         & {287}                                  & {303}                                  
\\ 
\midrule
     &                                & \textbf{1}              & 667 & \cellcolor[HTML]{DDEBF7}\textbf{2590} & 320                      & {{146}}     & 346 & {{135}} 
     \\ 
     &                                & \textbf{2}              & 735 & 2672& 322                      & {\cellcolor[HTML]{DDEBF7}\textbf{149}}     & 378 & {\cellcolor[HTML]{DDEBF7} \textbf{136}} 
     \\ 
     &                                & \textbf{25}             &{\cellcolor[HTML]{DDEBF7} \textbf{716}} & 2395& 321                      & 152                        & 162 & 138 \\ 
     &                                & \textbf{50}             & 726 & 2670& {\cellcolor[HTML]{DDEBF7}{\textbf{330}}} & {153} & 329 & 139 \\ 
     &                                & \textbf{105}            & 743 & 2760& 330                      & 155                        & 358 & 141 \\ 
     &                                & \textbf{215}            & 1285& 6122& 332                      & 159                        & 409 & 145 \\ 
     &                                & \textbf{430}            & 2398& 10330                                  & 348                      & 167                        & 683 & 154 \\ 
     &                                & \textbf{645}            & 3554& 10171                                  & 353                      & 177                        & 755 & 162 \\ 
\multirow{-9}{*}{\textbf{Traffic}}      & \multirow{-8}{*}{\textbf{861}} & \textbf{w/o PCA}        & 4626& {{19749}}               & 354                      & 173                        & \cellcolor[HTML]{DDEBF7}\textbf{625}   & 154 \\ 
\midrule
\end{tabular}
}
\caption{Runtime results of transformer forecasting models (w/o PCA) and PCA-enhanced transformers forecasting model (unit: second).}
\label{result_ela_time}
\end{table}
\subsection{Experimental Environment and Configuration}
All experiments are implemented on a platform with a single Nvidia TU02 GPU. We test all transformer-based models (refer to Table~\ref{tran_cate}), currently considered to be SOTA for time series forecasting. For each experiment, all models are configured with pre-selected default hyperparameters specific to each dataset for their best performance. 
The prediction length is set to 96 (based on the randomly selection) for all experiments as the long-term forecasting setting. For the models’ evaluation metrics, we use mean squared error (MSE) and mean absolute error (MAE). The code\footnote{\url{https://github.com/jingjing-unilu/PCA_Transformer}} in this paper is constructed based on the Time Series Library by~\cite{wu2023timesnet}. 
\subsection{Framework Construction}
Figure~\ref{structure_overview} represents the basic idea of the novel framework with various transformer-based forecasting models\footnote{They are all based on the vanilla transformer model.} serving as the backbone. The initial step of the experiments starts with loading the original dataset. The next step is to employ PCA and reduce a complete set of variables from $M$ to $P$. It is the process of dimensionality reduction. The \textit{target} variable is deliberately excluded from these steps to prevent information or data leakage. The following step is to partition the dataset resulting from PCA into training, testing and validation (depends on the model). The post-processed sub-dataset is then fed into transformer-based forecasting models for training and validation. The final step is using forecasting model to predict $F$ future values for the \textit{target} variable.\\
\textbf{- PCA Process.} 
In the PCA process, we utilize the randomized singular value decomposition (SVD) based PCA~\cite{scikit-learn,halko2011finding,szlam2014implementation} due to our large dataset. The input data of the PCA process is $H_{T\times M} = (x^{1}, x^{2}, ..., x^{M})\in \mathbb{R}^{T\times M}$ with $x^{m}\in\mathbb{R}^{T}$. The output data is $H_{T\times P}' = (c^{1}, c^{2}, ..., c^{P})\in \mathbb{R}^{T\times P}$. $P$ is principal components numbers of the PCA. The calculation steps are in the Algorithm~\ref{SVD_PCA}. 
\begin{algorithm}
\caption{SVD-based PCA}\label{SVD_PCA}
\small
\begin{algorithmic}[1]
\Procedure{PCA}{$H_{T\times M} = (x^{1}, x^{2}, ..., x^{M})\in \mathbb{R}^{T\times M}$ with $x^{m}\in\mathbb{R}^{T}$}
\State Center the dataset: $\tilde{H_{T\times M}} = $ $(\tilde{x^{1}}, \tilde{x^{2}}, ..., \tilde{x^{M}})$, where $\tilde{x^{m}} = (x^{m}-\frac{1}{T}\sum x^{m})$
\State Calculate the covariance matrix: $\tilde{CH_{T\times M}}=cov(\tilde{H_{T\times M}})$ 
\State Apply SVD: $\tilde{SH_{T\times M}}=svd(\tilde{CH_{T\times M}})$ 
\State Get top $P$ principal components: $H_{T\times P}'=\tilde{SH_{T\times P}}=(c^{1}, c^{2}, ..., c^{P})\in \mathbb{R}^{T\times P}$. 
\State \textbf{return} $H_{T\times P}'$
\EndProcedure
\end{algorithmic}
\end{algorithm}
\begin{table*}[!ht]
\centering
\resizebox{\textwidth}{!}{
\begin{tabular}{c|c|rr|rr|rr|rr|rr|rr}
\toprule
\multirow{2}{*}{\textbf{Dataset}} & \textbf{Variables} & \multicolumn{2}{c|}{\textbf{PCA+PatchTST}}  & \multicolumn{2}{c|}{\textbf{PCA+Crossformer}}  & \multicolumn{2}{c|}{\textbf{PCA+Autoformer}}   & \multicolumn{2}{c|}{\textbf{PCA+N.S. Trans.}}    & \multicolumn{2}{c|}{\textbf{PCA+iTransformer}}    & \multicolumn{2}{c}{\textbf{PCA+Transformer}}\\

  & \textbf{w/o Target}& \multicolumn{2}{c|}{(\bf MSE $\mid$ \bf Time)}  & \multicolumn{2}{c|}{(\bf MSE $\mid$ \bf Time)}   & \multicolumn{2}{c|}{(\bf MSE $\mid$ \bf Time)}    & \multicolumn{2}{c|}{(\bf MSE $\mid$ \bf Time)}     & \multicolumn{2}{c|}{(\bf MSE $\mid$ \bf Time)}     & \multicolumn{2}{c}{(\bf MSE $\mid$ \bf Time)} \\
\midrule
{ETTh1}       & 6 & {0.25\%} & 4.05\%  & {56.50\%} & 2.38\%  & {21.73\%} & 0.00\%  & {8.58\%}       & 33.33\%     & {0.43\%}  & 4.55\% & {0.00\%}  & 0.00\%    \\ 
{Weather}     & 20 & {0.00\%} & 0.00\%  & {57.64\%} & 30.77\% & {8.84\%}  & -0.58\% & {21.89\%}      & 0.48\%      & {8.29\%}  & -4.48\% & {0.00\%}  & 0.00\%   \\ 
{Electricity} & 320 &{2.14\%} & 88.68\% & {14.27\%} & 76.64\% & {0.00\%}  & 0.00\%  & {0.00\%}       & 0.00\%      & {53.20\%} & -32.40\% & {19.83\%} & 6.60\%   \\ 
{Traffic}     & 861 &{0.46\%} & 84.52\%  &{4.83\%}  & {86.9\% } & {17.75\%} & 6.78\%  & {17.70\%}      & 13.87\%     & {0.00\%}  & 0.00\%  & {5.21\%}  & 11.69\%   \\ 
\midrule
\multicolumn{2}{c|}{\textbf{Average Reduction}}    &0.71\%	& 44.32\%	& 33.31\%	& 49.17\%	& 12.08\%	& 1.55\%	& 12.04\%	& 11.92\%	& 15.48\%	& -8.08\%	& 6.26\%	& 4.57\% \\
\bottomrule
\end{tabular}
}
\caption{MSE and runtime reductions of PCA-enhanced transformer forecasting models. This table is the summary of the accuracy result Table~\ref{acc_result_PCA_tran} and corresponding runtime result Table~\ref{result_ela_time}. The baselines are corresponding original transformer models.}
\label{impro_result_PCA_tran}
\end{table*}
\\
\textbf{- Transformer-based Time Series Forecasting Models.} Figure~\ref{Tran_arch} depicts a vanilla transformer process with characteristic transformer components and architecture. Following the PCA process, we focus on a transformer-based time series forecasting process. The transformer process consists of three main components (attention, add \& norm, and feedforward) with two primary functions (encoder and decoder). According to the previous study~\cite{liu2023itransformer}, we can classify four transformer models (Crossformer, iTransformer, Autoformer, and PathTST) into modified components and architecture categories. Notice that non-stationary transformer belong to the same category of PatchTST, as well as our PCA-enhanced transformer framework.
%

\section{Experimental Results with Analysis}
\subsection{Model Perspective}
Table~\ref{impro_result_PCA_tran} summarises experiment results with PCA-enhanced and non-PCA-enhanced methods across all four datasets. Each entry in Table~\ref{impro_result_PCA_tran} represents the best-performing model from Table~\ref{acc_result_PCA_tran} and its runtime Table~\ref{result_ela_time} (lowest MSE with its runtime, bold with blue cell color). Table~\ref{impro_result_PCA_tran} shows that the proposed PCA-enhanced method exhibits significant reduction MSE and running time for all models except iTransformer. The results highlight that we can improve accuracy and efficiency by reducing input data dimensions. The entire process emphasizes the principle of ``less is more''. Based on Table~\ref{impro_result_PCA_tran}, we can summarize the experiment results as follows:  
\begin{enumerate}
\item Compared with non-PCA models, the \textbf{PCA+PathTST} model can reduce the average by \underline{0.71\%} on MSE and \underline{44.32\%} on runtime. In other words, PCA-enhanced model increases the model’s accuracy and time efficiency. For datasets of \texttt{Electricity} and \texttt{Traffic}, the \textbf{PCA+PatchTST} model can increase time efficiency \underline{88.68\%} and \underline{84.52\%}, respectively, while MSE can be reduced marginally by \underline{2.14\%} and \underline{0.46\%}. Similarly, the \textbf{PCA+Transformer} model improves performance when applying \texttt{Electricity} and \texttt{Traffic} datasets.
\item Regarding the \textbf{PCA+}\textbf{Non-stationary} \textbf{Transformer} model, we achieve a reduction of \underline{12.04\%} on the MSE and \underline{11.92\%} on the runtime. Likewise, \textbf{PCA+Autoformer} demonstrates similar results except for the weather dataset. 
\item Across all PCA-enhanced models, the \textbf{PCA+Cross-former} model has the best performance regarding the MSE reduction (\underline{33.31\%}) and the runtime decreasing (\underline{49.17\%}) compared to its non-PCA model. Table~\ref{impro_result_PCA_tran} shows that all PCA-enhanced models reduce the MSE (between \underline{0.71\%} and \underline{33.31\%}) and the runtime (between \underline{1.55\%} and \underline{49.17\%}) except the PCA+iTransformer model. 
\item Notice that the runtime of the \textbf{PCA+} \textbf{iTransformer} model goes in the opposite direction, increased by \underline{8.08\%} on average due to the model’s uniqueness of inverted design. Nevertheless, MSE is reduced by up to \underline{53.20\%} in the best case.
\end{enumerate}
\begin{figure}[!ht]
    \includegraphics[width=.23\textwidth]{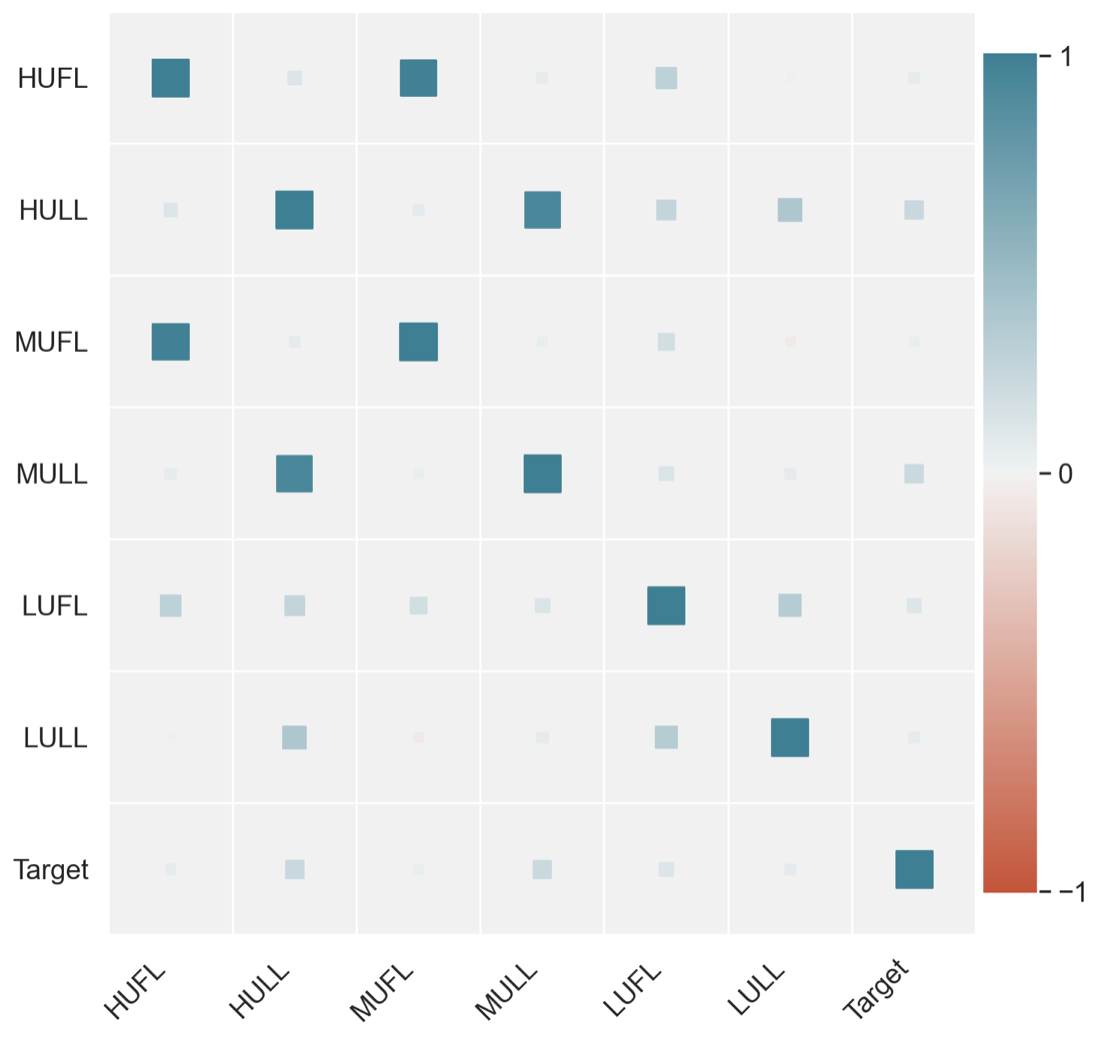}\hfill
    \includegraphics[width=.23\textwidth]{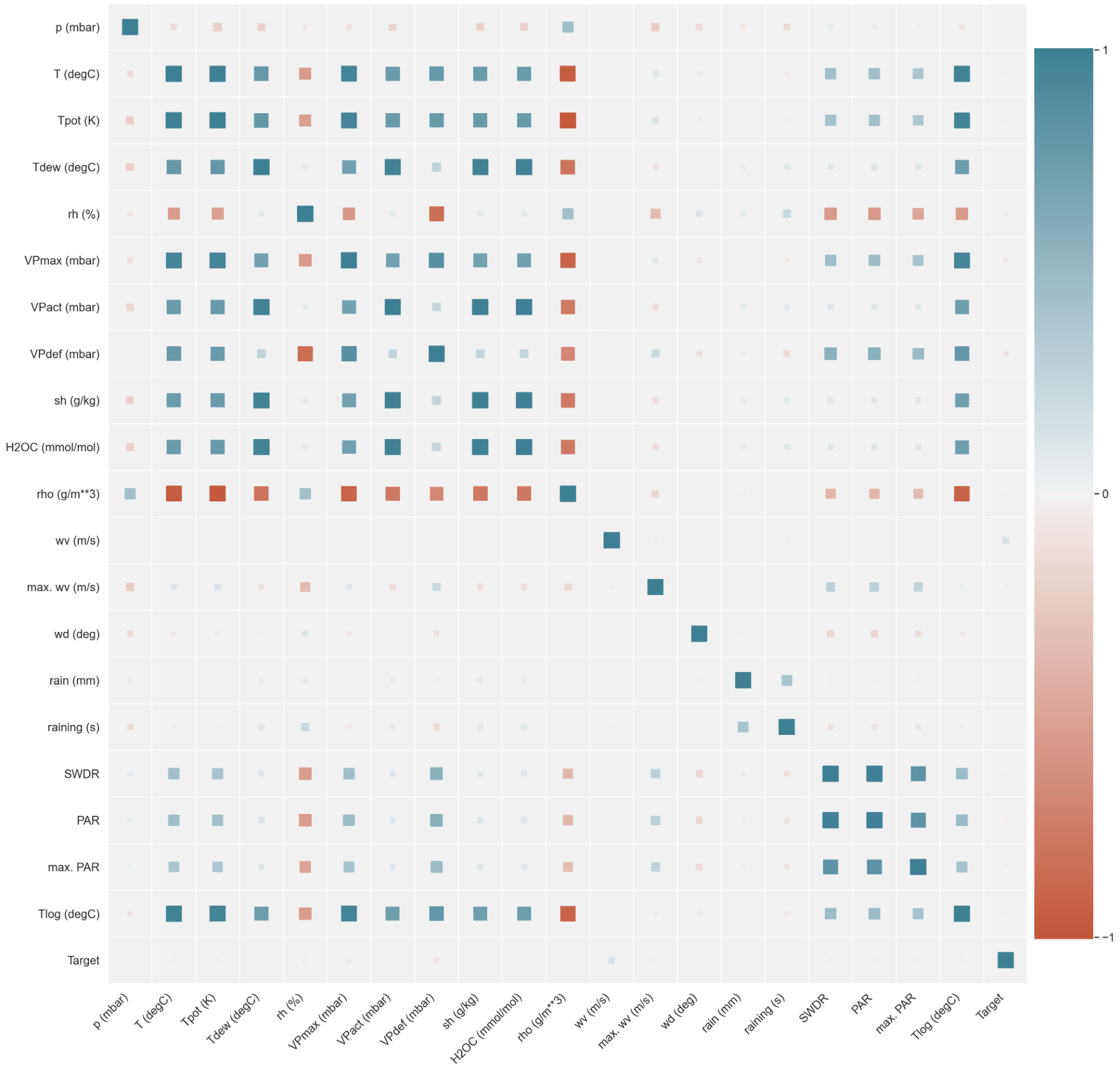}
    \caption{Left: PCC of ETTh1 dataset. Right: PCC of Weather dataset.}
    \label{PCC_ett_wea}
\end{figure}
\begin{figure}[!ht]
    \includegraphics[width=.22\textwidth]{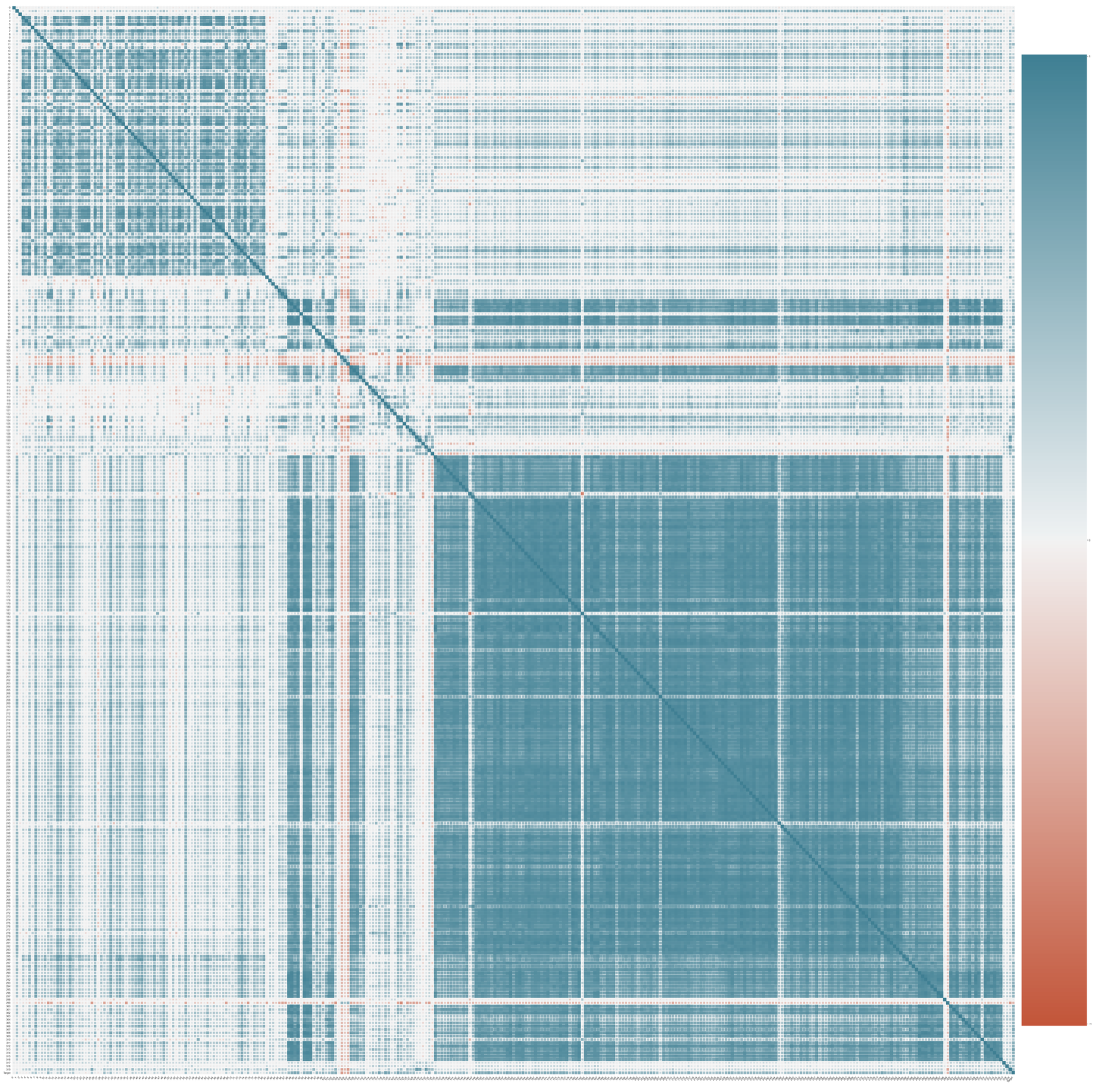}\hfill
    \includegraphics[width=.22\textwidth]{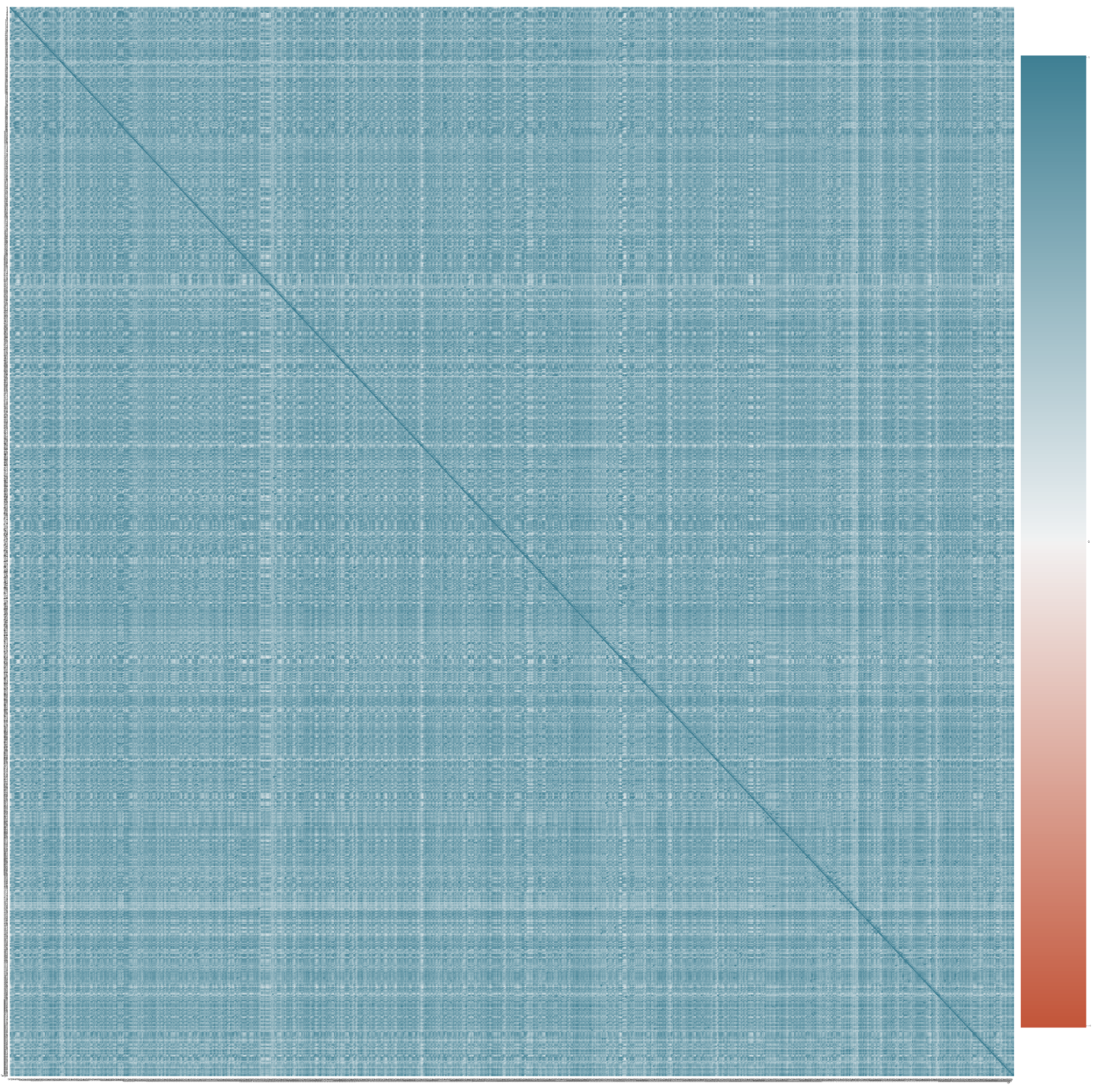}
    \caption{Left: PCC of Electricity dataset. Right: PCC of Traffic dataset.}
    \label{PCC_ele_tra}
\end{figure}
\begin{table*}[!ht]
\centering
\resizebox{0.95\linewidth}{!}{
\begin{tabular}{c|c|c|c|c|cc|rr|c}
\toprule
\multirow{2}{*}{{Dataset}}   & {Variables} & {{PCA}}  & {{Information}} & {{Dataset}} & \multicolumn{2}{c|}{{{Best Result}}}  & \multicolumn{2}{c|}{{{Best Reduction}}} & \multirow{2}{*}{{Model}}    
\\
      & {w/o Target} & {{Components}}  & {{{Kept (PCA)}}} & {{Ratio}} & \multicolumn{2}{c|}{( MSE $\mid$ Time)}     & \multicolumn{2}{c|}{( MSE $\mid$ Time)}     &  
\\
\midrule
{\textbf{ETTh1}}       & {{6}}         & {2}      & {70.1\%}   &{33.3\%}       & {{0.05561}} & {71}    & {{0.2\%}}  & {4.1\%}  & {PCA+Patch.}  
\\ 
\hline
{\textbf{Weather}}     & {{20}}       & {2}        & {63.1\%} & {10.0\%}      & {{0.00124}} & {653}   & {{8.3\%}}  & {-4.5\%}    & {PCA+iTrans.}
\\
\hline
{\textbf{Electricity}} & {{320}}               & {80}    & {{94.7\%}}     & {25.0\%}      & {{0.23528}} & {925}   & {{14.3\%}} & {76.6\%}  & {PCA+Crossf.}  
\\ \hline
{\textbf{Traffic}}     &{{861}}     & {1}      & {57.6\%}   & {0.1\%}    & {0.14789}    & {2590} & {4.8\%}       & {86.9\%}     & {PCA+Crossf.}         
\\ \midrule
\end{tabular}
}
\caption{Information Kept Ratio by PCA with the Best Performance Model (Lowest MSE on each Dataset). The baselines are corresponding original transformer models.}
\label{PCA_info_best_model}
\end{table*}
\begin{table}
\centering
\footnotesize
\resizebox{0.93\linewidth}{!}{
\begin{tabular}{c|c|c|c|c}
\toprule
\multirow{2}{*}{\textbf{Dataset}}   & {\textbf{Variables}} & {\textbf{PCA}}  & {\textbf{Information}} & {\textbf{Dataset}}
\\
      & {\textbf{w/o Target}} & {\textbf{Components}}  & {{\textbf{Kept (PCA)}}} & {\textbf{Ratio}}
\\
\midrule
      &         & {2}       & {70.1\%}    & {33.3\%}  
\\ 
    &         & {4}           & 99.6\%     & 66.7\%   
\\ 
\multirow{-3}{*}{\textbf{ETTh1}}       & \multirow{-3}{*}{{6}}         & {w/o PCA}    & -       & {100.0\%} 
\\ 
\midrule
    &         & {2}         & {63.1\%}   & {10.0\%}  
    \\ 
    &         & {5}         & {82.4\%}   & {25.0\%}  
    \\ 
    &         & {10}       & 98.9\%      & 50.0\%   
    \\ 
    &         & {15}      & 100.0\%    & 75.0\%     
    \\ 
\multirow{-5}{*}{\textbf{Weather}}     & \multirow{-5}{*}{{20}}       & {w/o PCA}    & -    & 100.0\%     \\ 
\midrule
    &         & {2}       & {{67.2\%}}  & {0.6\%}
    \\ 
    &         & {20}      & {87.4\%}     & 6.3\%  
    \\ 
    &         & {40}      & {91.2\%}  & 12.5\%  
    \\ 
    &         & {80}      & {{94.7\%}} & {25.0\%}  
    \\ 
    &         & {160}     & {97.7\%}   & 50.0\% 
    \\ 
    &         & {240}     & {99.2\%}    & 75.0\%    
    \\ 
\multirow{-7}{*}{\textbf{Electricity}} & \multirow{-7}{*}{{320}}       & {w/o PCA}    & - & 100.0\%  
\\ \midrule
    &         & {1}       & {57.6\%} & {0.1\%}  
    \\
    &         & {2}       & {71.2\%} & {0.2\%}    
    \\ 
    &         & {25}     & 83.5\%   & 2.9\% 
    \\ 
    &         & {50}     & 87.0\%  & 5.8\% 
    \\ 
    &         & {105}    & 91.2\%   & 12.2\%
    \\ 
    &         & {215}   & 95.3\%   & 25.0\%
    \\ 
    &         & {430}   & 98.5\%   & 49.9\%
    \\ 
    &         & {645}   & 99.6\%   & 74.9\%
    \\ 
\multirow{-8}{*}{\textbf{Traffic}}     & \multirow{-8}{*}{{861}}       & {w/o PCA} & -   & {100.0\%} 
\\ \midrule
\end{tabular}
}
\caption{Information Kept Ratio by PCA.}
\label{PCA_info}
\end{table}
\subsection{Dataset Perspective}
The essence of PCA is a linear, unsupervised transformation algorithm. It simplifies dimension reduction by identifying maximum variance in the data and incorporating new features~\cite{ghojogh2019feature}. This PCA characteristic leads us to employ the Pearson Correlation Coefficient (PCC) for evaluating linear correlations within datasets. Nevertheless, it is crucial to select correlation measurement approaches and customize them for particular dimension-reduction techniques. 

The basic logic of selecting the number of PCA components is based on the consideration of variable correlation. If all variables are independent, we should include all of them. Otherwise, we only select a few or even one. The correlation patterns in Figures~\ref{PCC_ett_wea} and~\ref{PCC_ele_tra} differ among the four datasets examined in this paper. The ETTh1 dataset with six variables shows two correlated variables, while the Weather dataset exhibits about 50\% of correlated variables in the right of the Figure~\ref{PCC_ett_wea}. Table~\ref{PCA_info} shows that if we employ a two-component PCA algorithm, the \texttt{ETTh1} dataset preserves \underline{70.1\%} of the information, the \texttt{Weather} dataset retains \underline{63.1\%}, the \texttt{Electricity} dataset maintains \underline{67.2\%}, and the \texttt{Traffic} dataset keeps \underline{71.2\%}. Nevertheless, we still selected the one-component-PCA method for the \texttt{Traffic} dataset experiment because this dataset has massive correlated variables (See the right of the Figure~\ref{PCC_ele_tra}). In order to achieve the best MSE performance, we select a different number of PCA components for different datasets. Table~\ref{PCA_info_best_model} summarises information on PCA components selection for the experiments. PCA-enhanced models consistently outperform their non-PCA counterparts across all datasets, highlighting the effectiveness of PCA in enhancing transformer-based forecasting models through dimensionality reduction on the input data. 

Table~\ref{PCA_info_best_model} also demonstrates that the PCA-enhanced model is particularly good for a dataset with a medium and large number of correlated variables. The more correlated variables are, the better PCA enhancement is. For example, the Electricity dataset illustrates this point, which can retain \underline{94.7\%} of the information after PCA enhancement, achieving MSE reduction by \underline{14.3\%} and runtime reduction by \underline{76.6\%}. Also, the Traffic dataset’s variables are highly correlated. The significant number of correlated variables means that all evaluation metrics benefit from PCA process: information preservation rate with one-component PCA is \underline{57.6\%}, while MSE is reduced by \underline{4.8\%} and runtime is decreased by \underline{86.9\%}.

\section{Conclusion and Future Works}
We present a novel forecasting framework that leverages Principal Component Analysis (PCA) to enhance the performance of transformer time series forecasting models. This study conducted experiments across five SOTA transformer models with four diverse real-world datasets. The subsequent analyses provide insights from both model and data perspectives. The results underscore the significant improvements achieved by the PCA-enhanced transformer forecasting framework across all contexts. This framework illuminates the substantial potential of dimension reduction algorithms in terms of both the effectiveness and efficiency of transformer-based forecasting models. We argue less is more. Nonetheless, this study primarily focuses on compressing datasets from a variable perspective. 

PCA-enhanced solution is a preliminary examination for dimension reduction in transformer models. Future research should extend its boundary to the temporal compression of datasets. Furthermore, it is may useful to consider other dimension reduction methods, such as linear discriminant analysis (LDA), Independent Component Analysis (ICA), GrandPrix, Zero-Inflation Modulation Analysis(ZIFA), t-distributed stochastic neighbour edging (t-SNE) etc. We hope that this study will serve as a stepping stone to stimulate future work on transformer-based time series forecasting.

\section{Acknowledgments}
This work was funded by the Luxembourg National Research Fund (Fonds National de la Recherche - FNR), Grant ID 15748747 and Grant ID C21/IS/16221483/CBD.

\bigskip

\bibliography{aaai24}

\end{document}